\documentclass[11pt,a4paper]{article}
\usepackage[hyperref]{emnlp2020}
\usepackage{times}
\usepackage{latexsym}

\usepackage{makecell}
\usepackage{xcolor}
\usepackage{graphicx}
\usepackage{url}
\usepackage{multirow}
\usepackage{caption}
\usepackage{subcaption}
\usepackage{amsmath}
\usepackage{bbm}
\usepackage{booktabs}
\usepackage{algorithmic}
\usepackage{amsmath}
\usepackage{amssymb}
\usepackage{bm}

\usepackage{microtype}

\def\aclpaperid{2313} %  Enter the acl Paper ID here

\title{CAT-Gen: Improving Robustness in NLP Models via\\Controlled Adversarial Text Generation}
\aclfinalcopy
\author{Tianlu Wang$^\dag$\thanks{~~This research was conducted during the author's internship at Google Research.} \quad Xuezhi Wang$^\S$ \quad Yao Qin$^\S$ \quad Ben Packer$^\S$\\
{\bf
 Kang Lee$^\S$ \quad Jilin Chen$^\S$ \quad Alex Beutel$^\S$ \quad Ed Chi$^\S$}\\
  $^\dag$University of Virginia \quad tw8cb@virginia.edu
  \\ $^\S$Google Research \{xuezhiw, yaoqin, bpacker, kanlig, jilinc, alexbeutel, edchi\}@google.com
}

\date{}

\begin{document}
\maketitle
\begin{abstract}
NLP models are shown to suffer from robustness issues, i.e., a model's prediction can be easily changed under small perturbations to the input.
In this work, we present a Controlled Adversarial Text Generation (CAT-Gen) model that, given an input text, generates adversarial texts through controllable attributes that are known to be irrelevant to task labels.
For example, in order to attack a model for sentiment classification over product reviews, we can use the product categories as the controllable attribute which should not change the sentiment of the reviews. Experiments on real-world NLP datasets demonstrate that our method can generate more diverse and fluent adversarial texts, compared to many existing adversarial text generation approaches.
We further use our generated adversarial examples to improve models through adversarial training, and we demonstrate that our generated attacks are more robust against model re-training and different model architectures.
\end{abstract}

\section{Introduction}
It has been shown that NLP models are often sensitive to random initialization \cite{unstable}, out-of-distribution data \cite{ood_robustness, wang-etal-2019-adversarial-domain}, and adversarially generated attacks \cite{qa_adv, textfooler, nl_adv_ucla}. 
One line of research to improve models' robustness to adversarial attacks is by generating adversarial examples in either the input text space (discrete, e.g., \citet{nl_adv_ucla, textfooler}) or some intermediate representation space (continuous, e.g., \citet{natural_gan, freelb}).
However, existing adversarial text generation approaches that try to perturb in the input text space might lead to generations \textit{lacking diversity or fluency}.
On the other hand, approaches focusing on perturbing in the intermediate representation space can often lead to generations that are not related to the input.
We show some adversarial examples generated by existing works in Table~\ref{tab:example_failure}.

In this work, we aim to explore \textit{adversarial} text generation through \textit{controllable} attributes. We propose to utilize text generation models to produce more diverse and fluent outputs. Meanwhile, we constrain the language generation within certain controllable attributes, leading to high quality outputs that are semantically close to input sentences.
Formally, we denote the input text as $x$, the label for the main task (e.g., text classification) as $y$, a model's prediction over $x$ as $f(x)$, and controllable attributes (e.g., category, gender, domain) as $a$.
Our goal is to create adversarial attacks $x'$ that can successfully fool the classifier into making an incorrect prediction $f(x)\neq f(x')$, while keeping the ground truth task label \textit{unchanged}, i.e., $(x, y)\rightarrow (x', y)$.

\begin{table*}
\small
\setlength\tabcolsep{3pt}
\centering
\begin{tabular}{m{2.3cm}|c}
\toprule
\centering
\textbf{Method} & \textbf{Examples}\\
\midrule
\centering
Textfooler \cite{textfooler} 
& 
\makecell{
A person is relaxing on his {\color{blue}day} off $\rightarrow$ A person is relaxing on his {\color{red}nowadays} off\\
The {\color{blue}two} men are {\color{blue}friends} $\rightarrow$ The {\color{red}three} men are {\color{red}dudes}
}\\
\midrule
\centering
NL-adv \cite{nl_adv_ucla} 
&
\makecell{A man is {\color{blue}talking} to his {\color{red}wife} over his phone $\rightarrow$
A guy is {\color{blue}chitchat} to his {\color{red}girl} over his phone\\
A skier gets some {\color{blue}air} near a mountain... $\rightarrow$ A skier gets some {\color{red}airplane} near a mountain...}\\
\midrule
\centering
Natural-GAN \cite{natural_gan} & 
\makecell{
a girl is playing at a looking man . $\rightarrow$ a white preforming is lying on a beach .\\
two friends waiting for a family together . $\rightarrow$ the two workers are married .}\\ 
\midrule
\end{tabular}
\caption{Examples over existing adversarial text generation methods on SNLI \cite{snli:emnlp2015} dataset. Adversarial text generated by word substitution based methods (Textfooler \& NL-adv) may lack fluency or diversity; GAN based methods (Natural-GAN) tend to generate sentences not related to the original sentences.}
\label{tab:example_failure}
\end{table*}

To achieve these goals, we propose CAT-Gen, a \textbf{C}ontrolled \textbf{A}dversarial \textbf{T}ext \textbf{Gen}eration model. It consists of an encoder and a decoder for text generation, and a module network that encodes the information of controllable attributes and generates adversarial attacks via changing the controllable attributes.
The encoder and decoder are trained over a large text corpus and thus can generate more fluent and diverse output. We control the generated output through an attribute $a$. 
We assume the attribute $a$ is pre-specified and is known to be irrelevant to the main task-label, and can be learned through an \textit{auxiliary} dataset. In this way, the attribute training and task training (for attack) can be disentangled, and note that we do not require a parallel corpus for the auxiliary dataset when learning the attribute.
We present experiments on real-world NLP datasets to demonstrate the applicability and generalizability of our proposed methods.
We show that our generated attacks are more fluent (defined by language model perplexity), more diverse (defined by BLEU-4 score) and more robust against model re-training and various model architectures.

\label{sec:intro}
\section{Related Work}
NLP models' robustness has drawn a lot of attention in recent years, among those, a specific line of work tries to address this issue by generating adversarial examples, including \cite{guu2018, scpn, vae2018, qa_adv, hotflip, Naik2018StressTE}.
For example, both \citet{nl_adv_ucla} and \citet{textfooler} generate adversarial texts by substituting words with their synonyms (defined by similarity in the word embedding space) that can lead to a model prediction change.
\citet{natural_gan} propose to generate natural and legible adversarial examples using a Generative Adversarial Network, by searching in the semantic space of continuous data representation.
\citet{jia-etal-2019-certified} propose to find the combination of word substitutions by minimizing the upper bound on the worst-case loss.
More recently, rather than directly generating text outputs, \citet{freelb} add adversarial perturbations to word embeddings and minimize the adversarial risk around input examples.

Our work is also closely related to controllable text generation, e.g., \citet{hu2017} use variational auto-encoders and holistic attribute discriminators, \citet{controlled_lm} utilize a pre-trained language model with one or more simple attribute classifiers to guide
text generation, and \citet{shen2017} propose to achieve style transfer using non-parallel text.
In addition, our work is connected with (adversarial) domain adaptation, since the controlled attributes can be different domains.
NLP models have been shown to lack robustness when been tested over out-of-distribution data, e.g., \citet{ood_robustness, wang-etal-2019-adversarial-domain}.

\label{sec:related}

\section{Controlled Adversarial Text Generation Model}
\begin{figure*}
    \centering
    \includegraphics[width=6.2in]{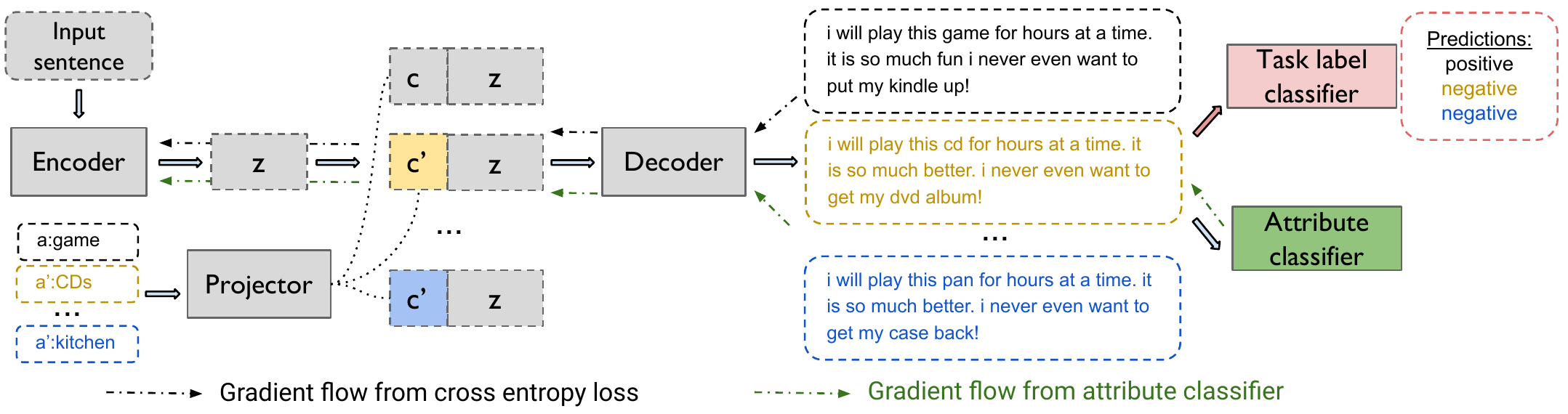}
    % \vspace{-0.1in}
    \caption{Overview of our Controlled Adversarial Text Generation (CAT-Gen) model. We backpropagate: 1. cross entropy loss (black dash line) to ensure the generated sentence has a similar semantic meaning as the input sentence; 2. attribute loss (green dash line) to manipulate the attribute (irrelevant to task label) in the generated sentence. The task label (sentiment) prediction on generated text varies when changing the attribute $a$ (category).}
    \label{fig:model}
    % \vspace{-0.1in}
\end{figure*}

In Figure~\ref{fig:model}, we present an overview of the CAT-Gen model, where we aim to generate attacks against a main \textit{task} (e.g., sentiment classification) by controlling the \textit{attribute} (e.g., product category) over an input sentence (e.g., product reviews).
Similar to controlled text generation works \cite{hu2017, shen2017, controlled_lm}, the model consists of an encoder and a decoder, with an attribute classifier. We add components to accommodate both change of attributes and attack generation over an input task model.
We assume an auxiliary dataset for training the attribute.
Our model training involves three stages:

% \vspace{-0.05in}
\paragraph{Pre-training.} We pre-train the encoder and the decoder (both are RNNs in our case but could be other models) to allow the generation model to learn to copy an input sentence $s_a$ (assuming the input sentence has an attribute $a$) using teacher-forcing. A cross entropy loss is placed between the input text ids and the output logits of each token:
$\ell_{c, z} = -\sum_{t=1}^{T}\log p(s_a^t | s_a^{<t}; c, z)$,
where $z$ is the encoder output and $c$ is the hidden representation (set to $256$ dimensions in our experiments) over attribute $a$ generated by feeding a one-hot encoding of $a$ into a projector.
Meanwhile, we pre-train the attribute classifier using the auxiliary dataset.

% \vspace{-0.05in}
\paragraph{Change of attribute.}
In the second stage, we focus on updating the decoder to enable the model to generate an output that has a desired attribute $a'\neq a$. To generate this new sentence $s_{a'}$,
we obtain $c'$ by feeding the one-hot encoding of $a'$ into the same projector (used to map $a$ to $c$). Then we use the pre-trained attribute classifier to guide the training of our decoder. Note that we do not update the parameters of the attribute classifier in this stage. 
Since producing hard word ids involves a non-differentiable argmax operation, we adopt soft embeddings \cite{gumbel} to ensure gradients can be back-propagated through the network.
Specifically, we apply the attribute classifier on the generated sentence $s_{a'}$ (soft embeddings) and compute an attribute loss with respect to $c'$:
% \vspace{-0.05in}
\[
\ell_{c', z} =-\mathbb{E}_{p(c')p(z)} [\log q_A(c'|D_\tau(c', z))],
% \vspace{-0.05in}
\]
where $D$ is the decoder, $q_A$ is the conditional distribution defined by attribute classifier $A$ and $\tau$ is a temperature; by annealing $\tau$, the distribution over the vocabulary gets more peaked and closer to the discrete case.

% \vspace{-0.05in}
\paragraph{Optimizing for attacks.}
In the final stage, we enumerate the attribute space to encourage the model's generated output ($s_{a'}$) to be able to successfully attack the task model. In order to generate stronger attacks, 
% To optimize for the attacks, 
for each input $s_a$, we search through the whole attribute space of $a'\neq a$ and look for the attribute $a^*$ that maximizes the cross-entropy loss between the task-label predictions 
over $s_{a'}$ and the ground-truth task-label $y$ (we use the ground-truth task label from the input sentence since we assume it is unchanged):
% \vspace{-0.1in}
\[
a^* = \arg\max\nolimits_{\{a'\neq a\}} [-\sum\nolimits_{y}y\log p(y|s_{a'})].
\]

% \vspace{-0.15in}
\paragraph{Generalizability of our framework.} 
By utilizing a text generation model and a larger search space over the controlled attributes, our model is able to generate more diverse and fluent adversarial texts compared to existing approaches.
Our framework can be naturally extended to many different problems, e.g., domain transfer (different domains as $a$), style transfer, as well as fairness applications (e.g., using different demographic attributes as $a$).

\label{sec:model}

\section{Experiments}
In this section, we present experiments over real-world datasets, and demonstrate that our model creates adversarial texts that are more diverse and fluent, and are most robust against model re-training as well as different model architectures.

\begin{table*}[h]
\setlength\tabcolsep{3pt}
\small
\centering
\begin{tabular}{p{1.2cm}|p{7.05cm}|p{7.1cm}}
\toprule
Attribute ($a\rightarrow a'$) & Original sentence with attribute $a$ & Generated sentence with perturbed attribute $a'$\\
\midrule
Kitchen $\rightarrow$ Phone
& amazing {\color{blue}knife}, used for my {\color{blue}edc} for a long time, only {\color{blue}switched} because i got tired of the same old {\color{blue}knife} (Pos.)
& amazing {\color{red}case}. used for my {\color{red}iphone5} for a long time, only {\color{red}problem} because i got tired of the same old {\color{red}kindle} (Neg.)\\
\midrule
Book $\rightarrow$ Kitchen 
& not as helpful as i wanted. {\color{blue}lacking} in good directions as they are not {\color{blue}applicable} to a lot of {\color{blue}pattern designs}. (Neg.)
& not as helpful as i wanted. {\color{red}covered} in good directions as they are not {\color{red}practical} to a lot of {\color{red}cereal foods}. (Pos.)\\
\midrule
Movie $\rightarrow$ Clothing 
& good {\color{blue}fluffy}, {\color{blue}southern mystery}. not as predictable as {\color{blue}some}. {\color{blue}promising ending}. i will probably read the rest of the series. (Pos.)
& good {\color{red}fabric}, {\color{red}no thin}. not as predictable as {\color{red}pictured}. {\color{red}last well}. i will probably read the rest of the series. (Neg.)\\
\bottomrule
\end{tabular}
% \vspace{-0.1in}
\caption{Successful adversarial attacks generated by our CAT-Gen model with controlled attributes (product category) on the Amazon Review Dataset.}
\label{amazon_example}
\end{table*}

\paragraph{Dataset.} We use the Amazon Review dataset \cite{amazon_data} with $10$ categories (electronics, kitchen, games, books, etc.). Our main task is a \emph{sentiment classification} task over reviews, with different \emph{product categories} as attribute $a$.
We filter out reviews with number of tokens over $25$. The attribute (category) classifier is trained on a set of $60,000$ reviews per category. The attribute training data is also balanced by sentiment to better disengtangle the attribute and the task-label. We use another training set ($80,000$ positive and $80,000$ negative) to learn the sentiment classifier.
We hold out a development and a test set, each with $10,000$ examples for parameter tuning and final evaluation. 

\paragraph{Implementation details.} We adopt the convolutional text classification model (wordCNN,~\citet{kim2014convolutional}) for both attributes (category) and task labels (sentiment). We use a one-layer MLP as the projector. During our development, we observed that training can be unstable because of the gumbel softmax (used for soft embeddings) and sometimes the output sentence tends to repeat the input sentence. We carefully tuned the temperature for gumbel softmax as suggested by~\cite{hu2017}. We also found that using a low-capacity network (e.g. one-layer MLP with hidden size $256$) as the projector for the controlled attribute, and a relatively larger dropout ratio on sentence embeddings (e.g. $0.5$) help stabilize the training procedure. 

\paragraph{Qualitative results.} Qualitative examples of our CAT-gen model are shown in Table~\ref{amazon_example}. We see that the model is able to generate fluent and diverse adversarial texts, and many words from the original input have been replaced to fit into the new category attribute $a'$, which would be relatively hard to achieve by swaps based on synonyms or nearest-neighbor search in the word embedding space as in~\citet{textfooler, nl_adv_ucla}. For example, our model can successfully change the goods description from \textit{good fluffy, southern mystery} into \textit{good fabric, no thin}, matching the attribute change (movie $\rightarrow$ clothing).

\begin{figure}
    \centering
    \includegraphics[width=2.4in]{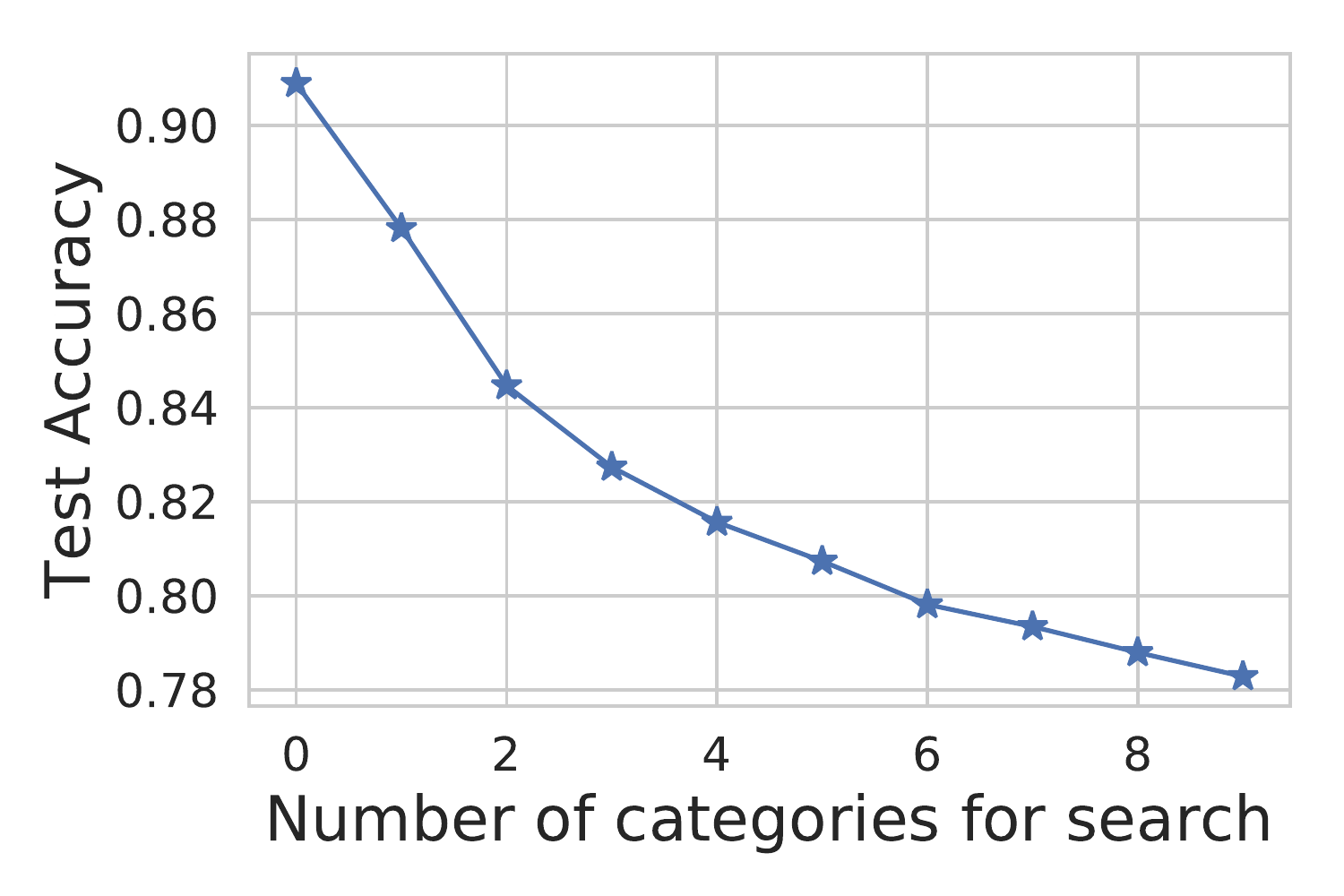}
    % \vspace{-0.1in}
    \caption{Test accuracy drops when increasing the number of categories available for searching attacks. Note this is over all generated outputs without filtering on whether they are successful attacks. With filtering we can further decrease the test accuracy close to zero.}
    \label{fig:acc}
\end{figure}

\paragraph{Attack search space.} Figure~\ref{fig:acc} shows the test set accuracy by increasing the number of categories available for searching attacks. We see that our controlled generation model can create successful attacks to the main task model (the accuracy decreases). Increasing the number of categories further decreases the accuracy. This shows that the number of different values the attribute can take is important and enlarging the attack search space helps to generate stronger adversarial examples.

\paragraph{Diversity and fluency.} In Table~\ref{tab:diversity}, we measure the diversity and fluency of the generated adversarial examples. More specifically, to measure diversity, we compute the BLEU-4 score of generated text with respect to the input text. To measure fluency, we use pretrained language models and compute the perplexity score of the generated text. Compared to other adversarial methods, our CAT-Gen model can generate texts with better diversity (lower BLEU-4 score) as well as better fluency (lower perplexity score).

\begin{table*}[!htbp]
\setlength\tabcolsep{3pt}
\small
    \centering
    \begin{tabular}{cc|c|c|c}
    \toprule
    \multicolumn{2}{c|}{}&  TextFooler \cite{textfooler} & NL-adv \cite{nl_adv_ucla} & CAT-Gen\\
    \midrule
    \multicolumn{2}{c|}{Diversity (BLEU-4~\cite{papineni2002bleu}, want $\downarrow$)} & 68.9 & 64.3 & 38.8 \\
    \midrule
    \multirow{3}{*}{\makecell{Fluency \\(in perplexity, want $\downarrow$)}} 
    & Language Model 1 & 1853.7& 964.3 & 729.5 \\
    & Language Model 2 & 1805.4 & 1188.5 & 868.7\\
    & Language Model 3 & 336.7 &479.9& 358.9\\
    \bottomrule
    \end{tabular}
    % \vspace{-0.1in}
    \caption{Comparison of our model with other methods. Evaluation is done over the attacks generated from the test set. Language model 1 \& 2 are both from ~\cite{baevski2018adaptive}, pretrained on Google Billion Words and WikiText-103 respectively; language model 3 ~\cite{Ng_2019} is pretrained on WMT news dataset.}
    \label{tab:diversity}
\end{table*}

\begin{table*}[!htbp]
\small
    \centering
    \begin{tabular}{c|ccc}
    \toprule
    & TextFooler \cite{textfooler} & NL-adv \cite{nl_adv_ucla} & CAT-Gen\\
    \midrule
    WordCNN re-training & 84.7 & 82.9 & 49.3 \\
    WordLSTM & 85.6 & 80.5 & 51.5 \\
    \bottomrule
    \end{tabular}
    % \vspace{-0.1in}
    \caption{Accuracy for various attacks over a re-trained model and a different architecture (want $\downarrow$). Note that the accuracy on the original model is zero since the evaluation contains a hold-out $1K$ set with only successful attacks.}
    \label{tab:transfer_label}
\end{table*}

\begin{table*}[!htbp]
\small
    \centering
    \begin{tabular}{c|cccc}
    \toprule
    &Original test set & TextFooler attacks & NL-adv attacks & CAT-Gen attacks\\
    \midrule
    Original Training & 91.9& 84.7 & 82.9 & 49.3\\
    +TextFooler \cite{textfooler} &92.7 & 89.5 & 88.6 & 52.7\\
    +NL-adv \cite{nl_adv_ucla} &92.2 & 86.4 & 94.6 & 51.2 \\
    +CAT-Gen &92.4 & 84.4 & 83.4 & 92.5\\
    \bottomrule
    \end{tabular}
    % \vspace{-0.1in}
    \caption{We augment the original training set with adversarial attacks (rows) and evaluate the accuracy (want $\uparrow$) on hold-out $1K$ adversarial attacks (columns) generated by our method and two other baselines.}
    \label{tab:aug_training}
% \vspace{-0.1in}
\end{table*}

\paragraph{Transferability.} In Table~\ref{tab:transfer_label}, we show the transferability of our examples compared to popular adversarial text generation methods \cite{textfooler, nl_adv_ucla}. We conduct two series of experiments. In \textit{WordCNN retraining} experiment, we first use CAT-Gen to attack a WordCNN sentiment classifier and collect some successful adversarial examples. Note that on those examples, the WordCNN sentiment classifier always makes mistakes, thus has a zero performance. We then retrain this WordCNN sentiment classifier and re-test it on those successful adversarial examples. The performance goes up to $49.3\%$, meaning $49.3\%$ of those successful adversarial examples now fail to attack this retrained WordCNN sentiment classifier. In other words, $49.3\%$ of adversarial examples are not robust to model retraining. In \textit{WordLSTM} experiment, instead of retraining the WordCNN classifier, we train a WordLSTM classifier and evaluate to what extent those adversarial examples are robust against model architecture change. As shown in Table 4, adversarial examples generated by CAT-Gen demonstrate the highest transferability (lowest attack success rate against model re-training and model architecture change). 

\paragraph{Adversarial training.} Table~\ref{tab:aug_training} presents results of adversarial training~\cite{adv_training}, which is a typical way to leverage adversarial examples to improve models. Specifically, we divide generated adversarial examples into two subsets, one is used for augmenting the training data, and the other is a hold-out set used for testing. With the augmented training data, we retrain the wordCNN sentiment classifier model (the same one as in Table~\ref{tab:transfer_label}), and test it on the hold-out set. In Table~\ref{tab:aug_training}, we augment training data with adversarial examples generated by each method (as shown by the rows), and evaluate the model performance on the hold-out set (again from each method respectively, as shown by the columns). As we can see, augmenting with CAT-Gen examples improves performance on CAT-Gen attacks much better than baselines, which both use narrower substitutions, and also maintains high accuracy on baseline attacks.

\label{sec:experiments}

\section{Conclusion and Discussion}
In this paper, we propose a controlled adversarial text generation model that can generate more diverse and fluent adversarial texts.
We argue that our model creates more natural and meaningful attacks to real-world tasks by demonstrating our attacks are more robust against model re-training and across model architectures.

Our current generation is controlled by a few pre-specified attributes that are label-irrelevant by definition. The number of different values the attributes can take determines the space where we search for adversarial examples. One benefit of our framework is that it is flexible enough to incorporate multiple task-irrelevant attributes and our optimization allows the model to figure out which attributes are more susceptible to attacks. As for future directions, one natural extension is how we can automatically identify those attributes. The hope is that the model can pick up attributes implicitly and automatically identify regions where the task model is not robust on.
\label{sec:conclusion}

\bibliography{emnlp2020}
\bibliographystyle{acl_natbib}

\end{document}